\documentclass{article}

\usepackage[final]{neurips_2018}

\usepackage[utf8]{inputenc} 
\usepackage[T1]{fontenc}    
\usepackage{hyperref}       
\usepackage{url}            
\usepackage{booktabs}       
\usepackage{amsfonts}       
\usepackage{nicefrac}       
\usepackage{microtype}      

\usepackage{amsmath,amssymb,amsfonts}
\usepackage{algorithmic}
\usepackage{graphicx}
\usepackage{textcomp}
\usepackage{xcolor}
\usepackage{float}
\usepackage[caption = false]{subfig}
\usepackage{natbib}

\newcommand{\R}{\mathbb{R}}

\title{Benchmarking and Optimization of \\ Gradient Boosting Decision Tree Algorithms}

\author{
  Andreea Anghel, Nikolaos Papandreou, Thomas Parnell \\
  \textbf{Alessandro de Palma, Haralampos Pozidis}\\
  IBM Research -- Zurich, R{\"u}schlikon, Switzerland \\
  \texttt{\{aan,npo,tpa,les,hap\}@zurich.ibm.com}
}

\begin{document}

\maketitle

\begin{abstract}
Gradient boosting decision trees (GBDTs) have seen widespread adoption in academia, industry and competitive data science due to their state-of-the-art performance in many machine learning tasks. 
One relative downside to these models is the large number of hyper-parameters that they expose to the end-user.
To maximize the predictive power of GBDT models, one must either manually tune the hyper-parameters, or utilize automated techniques such as those based on Bayesian optimization. 
Both of these approaches are time-consuming since they involve repeatably training the model for different sets of hyper-parameters.
A number of software GBDT packages have started to offer GPU acceleration which can help to alleviate this problem. 
In this paper, we consider three such packages: XGBoost, LightGBM and Catboost. 
Firstly, we evaluate the performance of the GPU acceleration provided by these packages using large-scale datasets with varying shapes, sparsities and learning tasks.
Then, we compare the packages in the context of hyper-parameter optimization, both in terms of how quickly each 
package converges to a good validation score, and in terms of generalization performance.
\end{abstract}

\section{Introduction}

Many powerful techniques in machine learning construct a \textit{strong learner} from a number of \textit{weak learners}. 
\textit{Bagging} combines the predictions of the weak learners, each using a different 
bootstrap sample of the training data set~\cite{breiman1996bagging}. 
\textit{Boosting}, an alternative approach, iteratively trains a sequence of weak learners, 
whereby the training examples for the next learner are weighted according to the success of the previously-constructed learners.
One of the widely-used boosting methods was AdaBoost \cite{freund1996experiments}. 
%
From a statistical perspective, AdaBoost iteratively assigns weights to training examples which has been shown 
to be equivalent to minimizing an exponential loss function \cite{buhlmann2007boosting}. The boosting technique proposed by~\cite{friedman}, 
known as \textit{gradient boosting}, generalizes this approach to minimize arbitrary loss functions. Rather than fit the next weak learner using weighted training examples, 
gradient boosting fits the next weak learner by performing regression on a function of the gradient vector of the 
loss function evaluated at the previous iteration. 

Recent gradient boosting techniques and their associated software packages have found wide success 
in academia and industry. XGBoost~\cite{KDD16_xgboost}, LightGBM~\cite{NIPS17_Ke}~\cite{arxiv17_Zhang} and Catboost~\cite{NIPS17_Anna} 
all use decision trees as the base weak learner and gradient boosting to iteratively fit a sequence of such trees. 
All packages belong to the family of gradient boosting decision trees (GBDTs) and expose a large number of hyper-parameters that the user
must select and tune. 
This tuning can either by done by hand, which can become a tedious and time-consuming task, or one 
can utilize techniques such as Bayesian hyper-parameter optimization (HPO).
While Bayesian optimization can automate the process of tuning the hyper-parameters, 
it still requires repeatedly training of models with different configurations which, for large datasets, can take a long time. 

In recent years, GPUs have seen widespread adoption in both on-premise and cloud computing environments. 
All three of the aforementioned GBDT packages are actively developing GPU-accelerated training routines, 
and all employ different algorithmic tricks to improve performance.
Clearly, if we can train each individual model more quickly, we can also accelerate the process of hyper-parameter tuning.
In this study, we present a comparison of the GPU-accelerated training functionality currently on 
offer in GBDT packages in the context of hyper-parameter optimization. Specifically, we will 
address the following three questions:
\begin{enumerate}
	\item How much acceleration can be expected when using GPU-based training routines?
	\item How well does this GPU-acceleration translate to reduced time-to-solution in the context of Bayesian hyper-parameter optimization?
	\item How well do the resulting models generalize to unseen data?
\end{enumerate}

The aim of our work is to address the three questions above by taking a rigorous experimental approach. We study the three aforementioned 
GBDT frameworks and evaluate their performance on four large-scale datasets with significantly different 
characteristics. 

\textbf{Related work.}
To the best of our knowledge, our paper is the first attempt to compare the GPU-acceleration provided by GBDT frameworks in the context 
of Bayesian hyper-parameter optimization for a diverse set of datasets and learning tasks. However, there have been a number of previous benchmarking efforts.
In ~\cite{gbdt-related-work-1} and~\cite{gbdt-related-work-2} the authors report the performance without 
using GPU acceleration and employ a grid search to tune the hyper-parameters. 
The authors of ~\cite{gbdt-related-work-3} report the GPU 
performance, but only for XGBoost and LightGBM and for a fixed set of hyper-parameters and a single dataset. 
~\cite{gbdt-related-work-4} 
compares the CPU implementations of LightGBM and XGBoost for binary classification and ranking tasks and reports the GPU performance only for LightGBM. A fixed 
set of hyper-parameters was used.
~\cite{NIPS17_Anna} reports accuracy results for datasets with categorical features showing the superiority of Catboost. The paper analyzes the Catboost GPU speedup improvement over its CPU implementation. The authors also acknowledge that it is very challenging to compare frameworks without hyper-parameter tuning, and opt to compare the three frameworks by hand-tuning parameters so as to achieve a similar level of accuracy. 

The paper is structured as follows.
In Section \ref{sec:GBTA} we review the GBDT algorithms. Section \ref{sec:framework} presents the frameworks used for hyper-parameter exploration. In Section \ref{sec:experimental_setup} we describe the experimental setup and the datasets. Section \ref{sec:results-hpo} first presents the GPU speedup results of the 3 GBDT algorithms (question 1). We 
also study the algorithms in the context of Bayesian HPO and compare them by runtime to discover the optimal hyper-parameters (questions 2 and 3). We conclude in Section \ref{sec:conclusion}. 
\vspace{-0.3cm}

\section{Gradient Boosting Decision Tree Algorithms}
\label{sec:GBTA}

Gradient boosting trees 
are tree ensemble methods that build a decision tree learner at a time by 
fitting the gradients of the residuals of the previously constructed tree learners.
Let $D$=$\{(\boldsymbol{x}_{i}, y_{i})\ |\ i\in\{1,..,n\}, \boldsymbol{x}_{i}\in \R^{m}, y_{i}\in \R\}$ be our dataset with $n$ examples and $m$ features. 
Given an ensemble of K trees, the predicted outcome $\hat{y(\boldsymbol{x})^K}$ for an input $\boldsymbol{x}$
is given by the sum of the values predicted by the individual K trees, $\hat{y(\boldsymbol{x})^K}=\sum_{i=1}^{K} f_{i}(\boldsymbol{x})$, where $f_{i}$ is the output of the $i$th regression tree of the K-tree ensemble. 
To build the ($K+1$)-th tree, the method minimizes a regularized objective function 
$L=\sum_{i=1}^{n} \ell(y_{i},\hat{y_i}^{K} + f_{K+1}(\boldsymbol{x}_i)) + \Theta(f_{K+1})$, where  $\ell(y_i,\hat{y_{i}}^{K} + f_{K+1}(\boldsymbol{x}_i))$ 
depends on the first (and possibly second) moments of the strong learner loss function $l(y_i, \hat{y_{i}}^{K})$, and $\Theta$ is 
a regularization function that penalizes the complexity of the ($K+1$)-th tree and controls the over-fitting. 

To build a tree the method starts from a single node and iteratively adds branches to the tree until a criterion is met. For each leaf branches are added so as to maximize the loss reduction after the split (\emph{gain function}). Iterating over all possible splits at each leaf is the bottleneck in training decision tree-based methods. The GBDT algorithms in this paper tackle the splitting task in various ways. 

XGBoost~\cite{KDD16_xgboost} proposes techniques for split finding and shows performance results for CPU training only. Two split methods are proposed: exact and approximate. The exact method iterates at each leaf over all the possible feature splits and selects the one that maximizes the gain. While very precise, this technique is slow. Thus an approximate algorithm \emph{hist} is proposed to calculate the split candidates according to percentiles of features distributions (histograms). The percentiles can be decided globally at the beginning of training or locally for each leaf. 

Two GPU implementations have been devised for approximate split finding.~\cite{PJCS17_Mitchel} proposes an implementation \emph{gpu\_hist} for XGBoost and~\cite{arxiv17_Zhang} for LightGBM. The XGBoost implementation builds a boosting tree without using the CPU thus reducing the CPU-GPU communication overhead during training. This method is fast, however, for large datasets such as~\cite{epsilon-challenge}, the GPU kernel fails due to GPU memory limitations. The LightGBM implementation uses the GPU only to build the feature histogram. The GPU kernel avoids using multi-scan and radix sort operations and reduces memory write conflicts when many threads are used for histogram construction.

The LightGBM library also includes an enhanced GBDT method based on sampling, gradient-based one-side sampling \emph{goss} \cite{NIPS17_Ke}, where input examples with small gradients are ignored during training. The paper proposes a CPU implementation, however the library allows us to use the \emph{goss} boosting type also in GPU. Thus we will use both the default \emph{gbdt} (GBDT) and \emph{goss} in our experiments.

Catboost is one of the most recent GBDT algorithms with both CPU and GPU implementations. It is a library that efficiently handles both categorical and numerical features. The GPU optimizations are similar to those employed by LightGBM. However, Catboost efficiently reduces the number of atomic operations when performing simultaneous computation of 32-bin histograms. It efficiently uses the shared memory per warp of threads and by using a good histogram layout it reduces write conflicts across threads. This translates in a high GPU occupancy.
\vspace{-0.3cm}

\section{Hyper-Parameter Exploration and Optimization}
\label{sec:framework}

To analyze the GPU efficiency of the GBDT algorithms we employ a distributed grid search framework. To evaluate how well the algorithms generalize to unseen data and to fine-tune the model parameters we use a HPO framework based on Bayesian optimization.

\subsection{Distributed Grid Search}

We implemented the distributed grid search using Apache Spark. First we load the training and validation datasets into memory on the master node. 
The datasets are then broadcast to each executor. Next, a resilient distributed dataset (RDD) is created consisting of all hyper-parameters sets
that should be evaluated. Each hyper-parameter partition is assigned to one executor. Each executor then proceeds to train the GBDT model for 
all of the hyper-parameters assigned to it. The models are evaluated on the validation set and the resulting evaluation metrics are collected 
on the master node.

One of the issues we encountered was how to correctly manage the GPUs from Spark. The system on 
which we deployed has 8 GPUs: 4 servers, each with 2 GPUs. 
We wanted to create a Spark RDD with 8 partitions, whereby the data from each partition is mapped to a unique GPU.
While this sounds straight-forward, there is a hidden complexity when one executes a distributed function call on an RDD, e.g., 
the Spark \texttt{mapPartitions} operation. 
By design, the executor handling each partition does not know on which host it is running and thus does not know which of the 2 GPU device IDs 
it can use. 
%
To overcome this challenge, we implement a file-system-based locking scheme. For each partition we 
clear a temporary directory of any previous lock files that may exist. Then for each partition we write a lock file containing the partition 
index, lock0 to lock7, into this same directory. After this stage, on each of the 4 servers, the temporary directory contains 2 lock 
files corresponding to the 2 partitions assigned to that server. Next, each partition reads a list of the lock files, sorts the filenames, and identifies its corresponding location in this sorted list. This way, each partition on a given server is assigned a number between 0 and 1 that corresponds to a unique GPU ID.

For the CPU experiments, we used 1 executor/server to fully utilize all the available CPU threads.


\subsection{Bayesian Optimization}\label{sec:bayes_opt}

Grid search requires extensive manual work to fine-tune the parameters and may be arbitrary in the choice of the grid points.
Hyper-parameter optimization (HPO) aims to automatize this tuning process. In Bayesian optimization, the prevalent
approach to HPO, the validation performance of the algorithm is treated as an unknown function of a hyper-parameter vector, 
$f(\boldsymbol{x})$, modeled through a probability distribution inferred from a regression (\emph{surrogate}) model.
An \emph{acquisition function} is used to select hyper-parameter sets that are both
informative and promising according to the learned model. For regression, popular
approaches use Gaussian Processes (GP) \cite{spearmint}.

We use HPO in our work to fine tune the models hyper-parameters and to 
compare the best generalization performance of the algorithms within a given parameter range.  
We used a GP model with Mat\'ern 5/2 kernel and expected improvement as the acquisition function~\cite{spearmint, FABOLAS}. 
As HPO library we chose to use~\cite{gpyopt2016} due to its flexibility in terms of both regression models and acquisition functions.
Bayesian optimization consists of iteratively evaluating new parameters according to an acquisition function, and updating the 
surrogate model with their results until a certain evaluation budget is exhausted. In our case, we stop after $150$ evaluations.
\vspace{-0.3cm}

\section{Experimental Setup}
\label{sec:experimental_setup}

\textbf{Datasets.} Table~\ref{tbl:dataset_characteristics} summarizes the datasets used in our experiments. 
Higgs and Epsilon are modeled 
as binary classification tasks. Microsoft (MSLR-WEB10K dataset) and Yahoo
consist of feature vectors extracted from query-URL pairs with relevance labels from 0 (irrelevant) to 4 (perfectly relevant).
Both problems can be modeled either as ranking or 5-class classification tasks. 
We chose the latter, as not all of the GDBT algorithms had ranking loss functions available at the time of the experiments.
For the GPU speedup results in Section \S\ref{sec:results-hpo}, we split the datasets into training and validation sets as shown in Table~\ref{tbl:dataset_characteristics}. 
Regarding the generalization performance in \S\ref{sec:results-hpo}, we employed the validation set as test set, and instead used a $25\%$ split of the training set as validation set. 

\begin{table}
	\setlength{\tabcolsep}{4pt}
	\renewcommand{\arraystretch}{1}
	\caption{Datasets characteristics.}
	\label{tbl:dataset_characteristics}
	\centering
	
	\begin{tabular}{|l|c|c|c|c|c|}  \hline
		& \multicolumn{2}{|c|}{Examples}    & Features    & Sparsity & Task \\ 
		\hline
		Dataset  	& Train 	& Validation	&      & Train		&  Train \\
		\hline
		Higgs~\cite{higgs}       & 10500000  & 500000 		& 28   & 7.89\% 	& Binary \\    
		Epsilon~\cite{epsilon-challenge}     & 400000    & 100000 		& 2000 & 0\%   		& Binary \\
		Microsoft~\cite{DBLP:journals/corr/QinL13}   & 723412    & 241521 		& 136  & 68.13\% 	& Multi-class \\
		Yahoo~\cite{yahoo-challenge}       & 473134    & 165660 		& 699  & 37.22\% 	& Multi-class \\
		\hline
	\end{tabular}
	
\end{table}


\textbf{Evaluation Metrics.} For Higgs and Epsilon, we calculate AUC-ROC, and for Microsoft and Yahoo the normalized discounted cumulative gain (NDCG)~\cite{pmlr-v30-Wang13} metric. We compute the expected relevance when class-probabilities are available (otherwise take the most probable relevance, for Catboost) and take the $10$ most relevant items to a query. We then compute the actual DCG 
and divide it by the DCG of the ideal item ordering to obtain NDCG-10.


\textbf{Hyper-Parameters Space.} In Table~\ref{tbl:par_range} we show the parameter set\footnote{We selected this parameter set to also run HPO in a reasonable amount of time, e.g. weeks.} used for both grid search and HPO experiments. The HPO ranges are wider than the extremes of the grid so as to avoid a priori assumptions, and to maximize the chances of including the optimum configuration. The GBDT algorithms share the same parameters, with the next exceptions: Catboost does not have feature fraction parameter, and LightGBM has two boosting types \emph{gbdt} and \emph{goss}. Finally, the number of leaves in LightGBM equals $2^\text{depth}$. We remark that smaller ranges were employed for Catboost on Epsilon for the number of iterations and the tree depth to fit in the GPU memory. 

\begin{table}
	\setlength{\tabcolsep}{4pt}
	\renewcommand{\arraystretch}{1}
	\caption{Hyper-Parameter Space for Grid-Search and HPO.}
	\label{tbl:par_range}
	\centering

	\begin{tabular}{|l|c|c|c|c|c|c|}  \hline
					& Iterations 		& Depth 	& Regularizer 	& Learn. Rate 	& Feature Frac. & Boosting \\ 
		\hline
		\multicolumn{7}{|c|}{Grid Search} \\
		\hline
		Catboost 	& 40,80,160,320,480 & 4,8,10,12 & 0,1,100 	& 0.1,0.3		& -- & -- \\
		\hline
		XGBoost 	& 40,80,160,320,480 & 4,8,10,12 & 0,1,100 	& 0.1,0.3		& 0.8,1.0 & -- \\
		\hline
		LightGBM 	& 40,80,160,320,480 & 4,8,10,12 & 0,1,100 	& 0.1,0.3		& 0.8,1.0 & gbdt,goss \\
		\hline
		\multicolumn{7}{|c|}{Hyper-Parameter Optimization (HPO)} \\
		\hline
		Catboost 	& [16,1000]*  &[2,14]* & $[10^{-2}, 10^{5}]$ 	& [0.01, 1]		& -- & -- \\
		\hline
		XGBoost 	& [16,1000] & [2, 14] & $[10^{-2}, 10^{5}]$ 	& [0.01,1]	& [0.01, 1] & -- \\
		\hline
		LightGBM 	& [16,1000] & [2, 14] & $[10^{-2}, 10^{5}]$ 	& [0.01,1]	& [0.01, 1] & gbdt,goss \\
		\hline
		
	\end{tabular}

	
\end{table}

\textbf{Hardware and Library Setup.} For all experiments we used 4 servers, each with 8-core Intel(R) Xeon(R) CPU E5-2630 v3, 64 GB RAM, and 2 NVIDIA GTX 1080 TI GPUs. Regarding the libraries, we used XGBoost 0.7, LightGBM 2.1.0, and Catboost 0.5.2.1.
\vspace{-0.3cm}

\section{GPU vs. CPU Performance and Hyper-Parameter Optimization Results}
\label{sec:results-hpo}

\begin{figure*}[t!]
	\centering
	\includegraphics[width=0.9\textwidth]{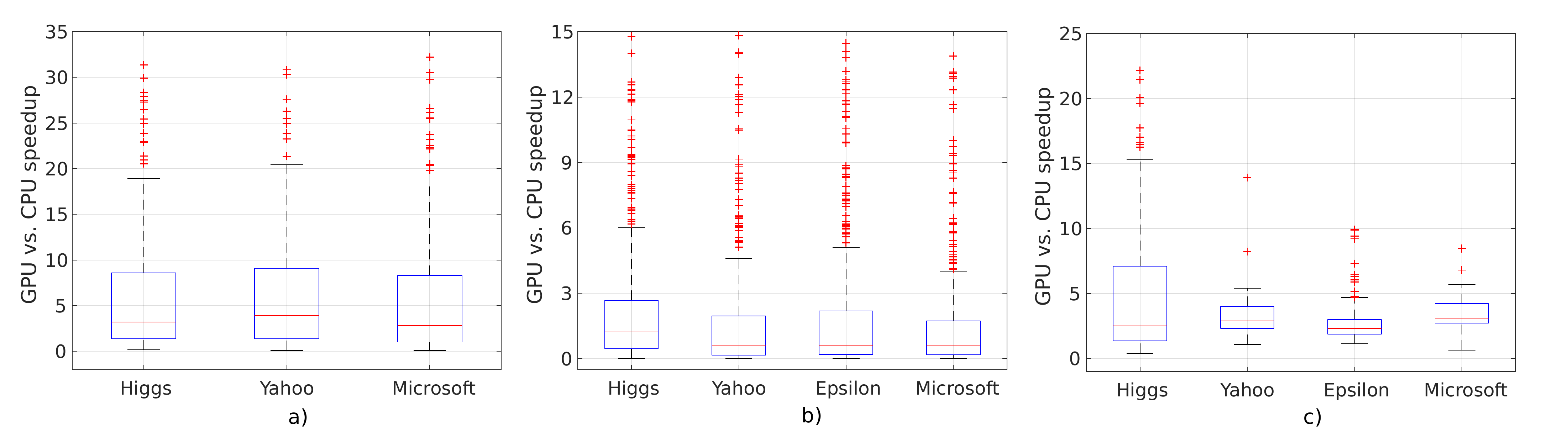}
	\vspace{-0.2cm}
	\caption{GPU vs. CPU performance across algorithms. (a) XGBoost, (b) LightGBM, (c) Catboost.} 
	\label{fig:time-to-accuracy}
\end{figure*}

We first compare the speedup of the GPU vs. CPU implementation per GBDT algorithm. We collect the speedup results for the grid search hyper-parameter combinations in Table~\ref{tbl:par_range}. Fig.~\ref{fig:time-to-accuracy}a)-c) show results for XGBoost, LightGBM and Catboost, respectively. XGBoost exhibits the highest speedup 7.26x on average and 3.32x at the median, followed by Catboost with an average speedup of 3.7x and 2.17x at the median. LightGBM has a lower average speedup of 2.35x and 0.75x at the median. For LightGBM we suspect that the \emph{goss} boosting routine is not yet fully optimized on GPU, thus negatively impacting the average speedup. Indeed, when excluding the \emph{goss} results, LightGBM exhibits a higher average speedup of 3.57x and 1.32x at the median.

Comparing the GPU training time of the GBDT frameworks for a fixed set of hyper-parameters is difficult since the 
hyper-parameters do not necessarily align and accuracy will not necessarily match. Thus, for a fair comparison, we 
study the performance in the context of HPO. Specifically, we want to know which GPU-accelerated GBDT algorithm can attain a good validation score 
in the shortest time when using the Bayesian optimization framework described in Section \ref{sec:bayes_opt}. 

\begin{figure*}[th]
	\centering
	\includegraphics[width=0.75\textwidth]{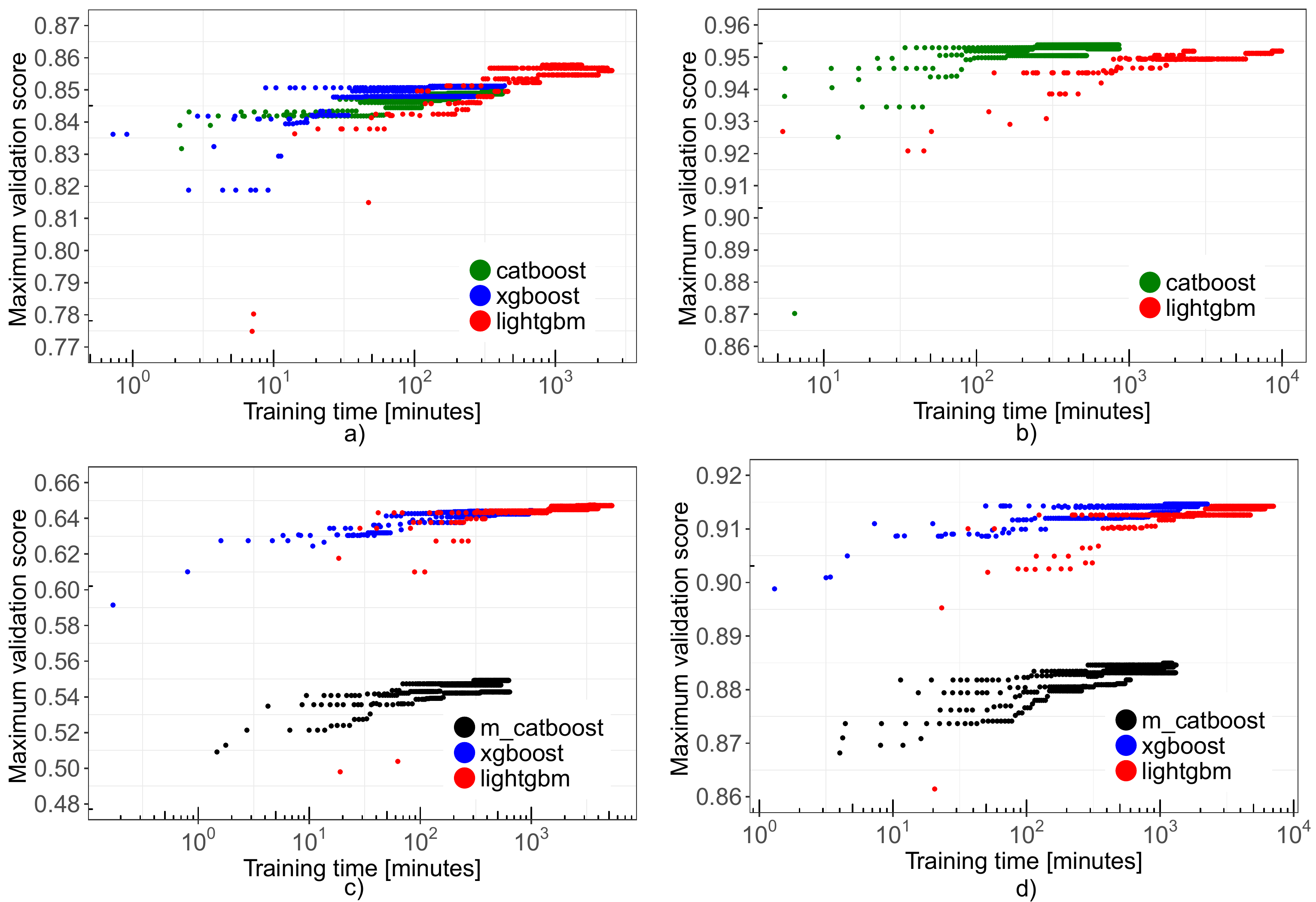}
	\vspace{-0.3cm}
	\caption{Max. validation score vs. total HPO runtime (a) Higgs (b) Epsilon (c) Microsoft (d) Yahoo. }
	\label{hpo-time-iterations}
\end{figure*}

In Figure~\ref{hpo-time-iterations} we plot the maximum validation score achieved vs. the total HPO runtime for all datasets. 
We plot three HPO runs per algorithm and dataset, each run with 150 iterations. In Table~\ref{tbl:best-hpo} we report the 
best validation score attained using HPO, as well as the score on the unseen test data for the resulting hyper-parameter configuration. 
As a baseline, we also include the scores achieved by a toy classifier that just predicts class labels according to their 
frequency in the training data. We would like to note that Catboost does not provide a multi-class classification loss function for GPU training\footnote{Starting with version 0.10.0, Catboost provides native GPU support for multi-class classification tasks.}. 
Thus, we implemented it as multiple one-vs-all binary classifications (\emph{m\_catboost}).

\textbf{Higgs - Figure~\ref{hpo-time-iterations}a)}. For the HIGGS dataset, we observe that XGBoost and Catboost can discover a configuration that leads to a good score much 
faster than LightGBM. This can be attributed to our previous observation, that XGBoost and Catboost can benefit much more 
significantly from GPU acceleration relative to LightGBM. However, while slower, LightGBM eventually finds a configuration that leads 
to a noticeably better validation score. Furthermore, this configuration generalizes and provides the best score on the test set. 

\textbf{Epsilon - Figure~\ref{hpo-time-iterations}b)}. XGBoost fails on Epsilon due to memory errors. While we again see that Catboost can evaluate configurations much faster, 
we do not observe LightGBM finding a better solution after more time based on this number of HPO iterations. In this case, the solution
that Catboost finds generalizes well and provides the best score on the test set. 

\textbf{Microsoft - Figure~\ref{hpo-time-iterations}c)}. For this dataset, which is a multi-class classification task, we again see that XGBoost can locate a good configuration significantly 
faster than LightGBM, but that after some time LightGBM is able to find a configuration that results in a higher validation score. However, in 
this case, the improvement in the validation score does not carry across to the test set, and the solution XGBoost found is the winner. We also observe that 
Catboost cannot locate a good configuration of hyper-parameters for this task.
We note that for all algorithms, including the baseline (naive classifier that predicts the labels according to their probability distribution in the training set), 
the validation score achieved is significantly higher than the corresponding test score. This effect is 
not caused by over-fitting the validation set, but is instead related to imbalance in the statistics between the validation and test set. While a stratified 
split was utilized to ensure that the distribution of class labels is preserved across train, validation and test sets, the NDCG score also depends on the statistics 
of the query IDs, which vary significantly.

\textbf{Yahoo - Figure~\ref{hpo-time-iterations}d).} As for the other multi-class classification task, we find that Catboost does not find a good solution. 
XGBoost is significantly faster, and can find a good configuration. Furthermore, the solution XGBoost finds also provides the best score on the test set. 

To summarize, we find that, while XGBoost can explore its space of hyper-parameters very fast, it does not always locate 
the configuration that results in the best score. While it clearly wins in both multi-class ranking tasks (Microsoft, Yahoo), for the 
Higgs dataset it loses to LightGBM, despite the latter being significantly slower.  Furthermore, for the Epsilon dataset 
XGBoost cannot be used due to memory limitations. Clearly, there is no one-size-fits-all solution to optimizing GBDT models, 
and we hope that these results may assist developers of the respective packages by identifying learning tasks for which there 
is still room for improvement.

\begin{table}[t]
	\setlength{\tabcolsep}{4pt}
	\renewcommand{\arraystretch}{1}
	\caption{Best test scores across algorithms and datasets.}
	\label{tbl:best-hpo}
	\centering
	
	\begin{tabular}{|l|c|c|c|c|c|c|c|c|}  \hline
		Dataset & \multicolumn{2}{|c|}{Baseline} & \multicolumn{2}{|c|}{XGBoost} & \multicolumn{2}{|c|}{LightGBM} & \multicolumn{2}{|c|}{Catboost} \\ 
		\hline
		& Test & Val & Test & Val & Test & Val & Test & Val  \\
		\hline
		
		Higgs       & 0.4996 & 0.5005 & 0.8353 & 0.8512 & \textcolor[rgb]{0,0,1}{\emph{0.8573}} & 0.8577 & 0.8498 & 0.8496 \\
		Epsilon     & 0.4976 & 0.5008 & - & - & 0.9513 & 0.9518 & \textcolor[rgb]{0,0,1}{\emph{0.9537}} & 0.9538 \\
		Microsoft   & 0.2251 & 0.3974 & \textcolor[rgb]{0,0,1}{\emph{0.4917}} & 0.6443 & 0.4871 & 0.6473 & 0.3782 & 0.5492\\
		Yahoo       & 0.5802 & 0.8106 &\textcolor[rgb]{0,0,1}{\emph{0.7983}} & 0.9146 & 0.7965 & 0.9142 & 0.7351 & 0.8849 \\
		
		\hline              
	\end{tabular}
	
\end{table} 

%
%

\section{Conclusions}
\label{sec:conclusion}

We have presented an in-depth experimental analysis of three state-of-the-art GBDT packages: XGBoost, LightGBM and Catboost. 
We have quantified the level of GPU acceleration that is currently provided by each of the packages, and 
evaluated how well this speedup translates into reduced time-to-accuracy in the context of Bayesian HPO. 
We have observed that, for a fixed set of hyper-parameters, XGBoost provides the largest reduction in training time 
when using a GPU relative to a CPU. Moreover, we observe that one is often able to utilize this speedup to 
converge to a good set of hyper-parameters in a short time. However, there are tasks for which LightGBM, albeit slower, can 
converge to a solution that generalizes better. Furthermore, for datasets with a large number of features, XGBoost 
cannot run due to memory limitations, and Catboost converges to a good solution in the shortest time. 
Therefore, while we observe interesting trends, there is still no clear winner in terms of time-to-solution 
across all datasets and learning tasks. The challenge of building a robust GPU-accelerated GBDT framework that 
excels in all scenarios is thus very much an open problem. 

\newpage {

	\footnotesize{
		Intel is a trademarks or registered trademarks of Intel Corporation or its subsidiaries in the United States and other countries.
		The Apache Software Foundation (ASF) owns all Apache-related trademarks, service marks, and graphic logos on behalf of our 
		Apache project communities, and the names of all Apache projects are trademarks of the ASF.
	}
	
	\small
	\bibliographystyle{unsrt}
	\bibliography{nips_2018_submission}

}
\end{document}